# Lyapunov Exponent as Physics-Informed Dense Reward: RL Discovery of Stabilization Beyond the Kapitza Pendulum

Slava Andrejev

We suggest using the Lyapunov characteristic exponent (LCE) as a dense reward signal for the reinforcement learning problem of stabilizing the inverted pendulum with vertical motion. With LCE, the agent not only successfully found the oscillatory motion known as the Kapitza pendulum but also damped the pendulum's pivoting, leaving it in a strictly upright position.

## 1 Introduction

One of the classic problems for testing reinforcement learning methods is the inverted pendulum problem. In the textbook formulation, a mass on a light, rigid stick is attached to a cart on a rail, with the mass pointing upward. An RL agent must find a strategy for cart movement to keep the pendulum upright.

Besides this straightforward way of balancing the pendulum with a horizontal motion, we can also set it upright with a vertical motion. If we oscillate the pivot point up and down fast enough, the pendulum will have an equilibrium point at the top. This type of pendulum, with an oscillating pivot, is called the Kapitza pendulum. To restore justice, Pyotr Kapitza didn't discover it; he found a brilliant way to describe its motion [1]. The fact that this pendulum has a stable equilibrium upright has been known before [2].

A fair question to ask is: if we apply the reinforcement learning methods we use for the inverted pendulum on a cart to a pendulum with the pivot moving up and down, will the AI agent find the Kapitza pendulum? In one attempt [3], the authors used the Deep Deterministic Policy Gradient (DDPG) method [4]. The agent failed to find oscillatory movement. It either let the pendulum fall most of the time at the end of an episode, or started accelerating downward without bound. The authors hypothesized that the reward sparsity was the reason for the failure. They used the squared angle with vertical as a reward signal. They didn't verify their hypothesis.

To improve their result, we suggest using the maximal Lyapunov Characteristic Exponent (LCE) [5] as a reward signal. From here on, we will omit "maximal". Whenever we write LCE, we will assume the maximal LCE. The LCE is an important parameter of a dynamical system. It measures how fast infinitesimally close trajectories in the phase space diverge. For the purpose of stabilization, a negative LCE indicates the system is stable. If the LCE is positive, infinitesimally small perturbations of the initial state grow and diverge chaotically. For the inverted pendulum, the LCE is positive. As a consequence, we cannot balance the inverted pendulum; we have to apply some form of control to keep the system upright.

If we compute an average LCE over a window, it will give the agent a direct signal about whether the system is stable over time. The squared angle with vertical does not carry this integral information. If the pendulum falls in the middle of an episode, we don't know how well we performed before the fall. We just get a large negative reward with no indication that oscillatory movement would have led to a stable system.

This work will explore whether an agent, armed with LCE as a reward signal, can discover inverted pendulum stabilization with vertical movement of the pivot point.

## 2 Motion Equations

Figure 1 shows the physical model of our pendulum. It is not difficult to show that the Lagrangian of the system is

$$L = \frac{m}{2}\left(\dot{x}_c^2 + \dot{y}_c^2 + l^2\dot{\varphi}^2 + 2l\dot{\varphi}\left(\dot{x}_c\cos\varphi + \dot{y}_c\sin\varphi\right)\right) - \\ - mgl\cos\varphi,$$

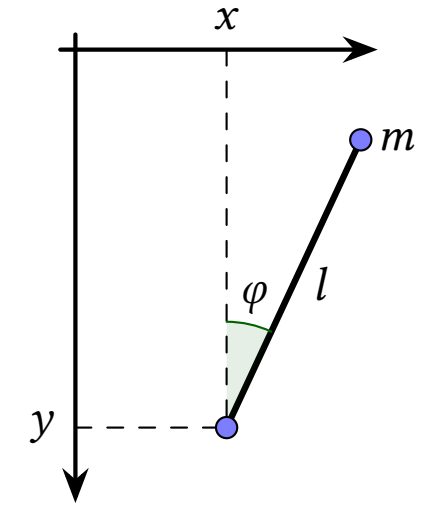

**Figure 1:** Pendulum model.

where $x_c$ and $y_c$ are the coordinates of the pivot point. Using the Euler–Lagrange equation for $\varphi$, we will get

$$\ddot{\varphi} + \frac{\ddot{x}_c}{l}\cos\varphi + \frac{\ddot{y}_c - g}{l}\sin\varphi = 0.$$

Since we have only vertical motion, $\ddot{x}_c = 0$, and $\ddot{y}_c$ will be our action (RL, not physical action), which we will denote as $a$. Finally,

$$\ddot{\varphi} + \frac{a - g}{l}\sin\varphi = 0. \tag{1}$$

For simplicity, we will measure acceleration in units of $g$ and distance in units of $l$. With this choice, the natural time scale becomes $\sqrt{l/g}$ seconds. (1) will take the following form

$$\ddot{\varphi} = (1 - a)\sin\varphi, \tag{2}$$

where $a(t) = \ddot{y}_c(t)$.

## 3 The Maximal LCE

We will follow Benettin et al. [6, 7] in briefly describing the LCE theory and the numerical method for computing the LCE.

We have a compact connected Riemannian manifold $M$ and a diffeomorphism $\Phi$ of $M$ onto itself. At each point $x \in M$, $T_xM$ is the tangent space of $M$ at point $x$. A flow $\{\Phi^t\}$ is a one-parameter family of diffeomorphisms satisfying $\Phi^{s+t} = \Phi^s \circ \Phi^t$, and $\Phi^0 = \mathbf{1}$. $d\Phi^t_x : T_xM \to T_{\Phi^t x}M$ is a linear map telling us how infinitesimally nearby points get stretched and rotated when we apply $\Phi$. Finally, $\{d\Phi^t_x\}$ is a family of linear maps.

For any nonzero vector $g \in T_xM$ one can define

$$\chi(g) = \limsup_{t\to\infty} \frac{1}{t} \ln \left\| d\Phi^t_x\, g \right\|. \tag{3}$$

This number is called the LCE of vector $g$ with respect to the family $d\Phi^t_x$. It can be shown [6] that $\{\chi(g)\}_{g\in T_xM}$ has at most $n = \dim T_xM$ different values. These values are $\nu_1 > \nu_2 > \ldots > \nu_s$, with $1 \leqslant s \leqslant n$.

All $g$ that produce $\nu_s$ in (3) form a linear space $L_s$. Those $g$ that produce $\nu_s$ and $\nu_{s-1}$ form a linear space $L_{s-1} \supset L_s$. We can continue this process until we reach $L_1$. Therefore, we have

$$L_s \subset L_{s-1} \subset \ldots \subset L_1 = T_xM,$$

and for a vector $v \in L_i \setminus L_{i+1}$

$$\nu_i = \lim_{t\to\infty} \frac{1}{t} \ln \left\| d\Phi^t_x\, v \right\|, \quad 1 \leqslant i \leqslant s. \tag{4}$$

If we randomly choose vector $g \in T_xM$ and plug it to (4), we will obtain the maximal LCE $\nu_1$ because $g$ has zero probability to end up in $L_2, \ldots, L_s$. This is the basis of computing the maximal LCE numerically as it is shown in [7].

For numeric computation, we have to take care of the exponential growth of the evolving tangent vector $d\Phi^t_x\, v$. To avoid floating point overflow, Benettin et al. [7] suggest periodically renormalizing the tangent vector after some fixed time $s$. That is, we choose a random $v \in T_xM$ with $\|v\| = 1$ and compute recursively

$$\begin{aligned} w_0 &= v \in T_xM, \\ \alpha_k &= \left\| d\Phi^s_{x_{(k-1)s}}\, w_{k-1} \right\|, \\ w_k &= \frac{d\Phi^s_{x_{(k-1)s}}\, w_{k-1}}{\alpha_k}. \end{aligned}$$

Finally,

$$\nu_1 = \lim_{k\to\infty} \frac{1}{ks} \sum_{i=1}^{k} \ln \alpha_i. \tag{5}$$

An ODE for $d\Phi^t_x$ can be derived using the Jacobian matrix. If we have a dynamic system $\dot{x} = f(x)$, then $\Phi^t_x$ is a solution that maps the initial condition $x(0) = x$ to $x(t)$. We can write this fact down as

$$\frac{d}{dt}\Phi^t_x = f\left(\Phi^t_x\right). \tag{6}$$

$d\Phi^t_x$ by definition is the spatial derivative of $\Phi^t_x$ with respect to the initial condition. If we apply the spatial derivative to (6) and use the chain rule, we will get an ODE for $d\Phi^t_x$

$$\frac{d}{dt}\left[d\Phi^t_x\right] = Df(x) \cdot d\Phi^t_x\,,$$

where $Df$ is the Jacobian matrix.

For the case of the inverted pendulum, $M$ consists of pairs $x = (\varphi, \dot{\varphi})$. From (1), the Jacobian matrix is

$$Df = \begin{bmatrix} 0 & 1 \\ \frac{1}{l}\left(g - a\left(t\right)\right)\cos\varphi & 0 \end{bmatrix}.$$

Similarly, $T_x M$ is a linear space of pairs $(\delta, \dot{\delta})$. Using the Jacobian above, we get

$$\ddot{\delta} = \frac{g - a\left(t\right)}{l}\delta\cos\varphi. \tag{7}$$

With the units we introduced earlier, this equation becomes

$$\ddot{\delta} = (1 - a)\,\delta\cos\varphi. \tag{8}$$

Instead of using a random initial vector $w_0 = (\delta_0, \dot{\delta}_0)$ for an LCE estimate in (5), we can directly derive an approximation of the direction of the maximum $\|w_0\|$ change. We set our pendulum at some small initial angle $\varphi_0$ and let it go. The solution of (2) for small angles is

$$\varphi \approx \varphi_0\cosh(t) \approx \varphi_0.$$

Since $\|w_0\| = 1$, we can write a solution of (8) as

$$\delta\left(t\right) = \cos\alpha_0\cosh(kt) + \frac{\sin\alpha_0}{k}\sinh(kt),$$

where

$$k = \sqrt{\cos\varphi_0},$$

and $\alpha_0$ is the initial direction of $w_0$. We need to find $\alpha_0$ that maximizes $\|(\delta, \dot{\delta})\|$.

$$\frac{d}{d\alpha_0}\left\|(\delta, \dot{\delta})\right\|^2 = \frac{1 - k^4}{k^2}\sin(2\alpha)\sinh^2(kt) + \frac{1 + k^2}{k}\sinh(2kt)\cos(2\alpha).$$

From which it follows that at

$$\alpha_0 = \frac{\pi}{2} - \frac{1}{2}\operatorname{atan}\left(\frac{2k\cosh(kt)}{\left(1 - k^2\right)\sinh(kt)}\right)$$

we will observe the maximum increase of $\|w_0\|$. $t$ in the above will be the time period over which we estimate the LCE.

## 4 LCE Solution For a Pendulum With an Oscillating Pivot

Let's verify that a solution for (2) and (8) with a constant angle $\varphi$ and negative LCE exists. We assume that $a(t)$ and $y_c(t)$ are fast-oscillating periodic functions of period $T$. Following [8], we further require that the changes in $\varphi$ are small, and we separate $\varphi$ into slow and fast parts,

$$\varphi = \varphi_s + \varphi_f. \tag{9}$$

Angle $\varphi_f$ is a periodic function and its value over one period of oscillation remains small. Angle $\varphi_s$ can have any value but during the same time $T$ it doesn't change much. If we average these values over the period $T$ and mark this average with a bar, we get

$$\overline{\varphi} \cong \varphi_s, \quad \overline{\varphi}_f \cong 0. \tag{10}$$

If we substitute (9) into (2), drop the slow $\ddot{\varphi}_s$, and expand $\sin\varphi$, we get

$$\ddot{\varphi}_f = (1 - a)\left(\sin\varphi_s + \varphi_f\cos\varphi_s\right).$$

Since $\varphi_f$ is small, we can leave out $\varphi_f\cos\varphi_s$, and the term $1 \cdot \sin\varphi_s$ is a slow drift, which belongs to $\varphi_s$. Therefore,

$$\ddot{\varphi}_f = -a\sin\varphi_s,$$

and

$$\varphi_f = -\eta\sin\varphi_s, \tag{11}$$

where $\eta(t)$ is the zero-mean shift of $y_c$, $\eta = y_c - \overline{y_c}$, due to the requirement $\overline{\varphi}_f = 0$ from (10). The next step is to use (11) in (2) and (9),

$$\ddot{\varphi}_s + \ddot{\varphi}_f = (1 - a)\left(\sin\varphi_s + \varphi_f\cos\varphi_s\right),$$

and average the result over one period $T$,

$$\ddot{\varphi}_s = \sin\varphi_s + \overline{a\eta}\cos\varphi_s\sin\varphi_s.$$

After integration by parts,

$$\ddot{\varphi}_s = \sin\varphi_s - \overline{\dot{\eta}^2}\cos\varphi_s\sin\varphi_s.$$

This gives a necessary condition for stabilizing the pendulum at a nonzero constant angle $\varphi_s$,

$$\overline{\dot{\eta}^2} = \cos^{-1}\varphi_s. \tag{12}$$

We can follow a similar technique for $\delta$ in (8). We separate it into slow and fast parts,

$$\delta = \delta_s + \delta_f.$$

Finding $\delta_f$ is very similar to finding $\varphi_f$,

$$\delta_f = -\eta\delta_s\cos\varphi_s.$$

The slow part is more convoluted,

$$\ddot{\delta}_s = \delta_s \cos\varphi_s + \delta_f \cos\varphi_s - a\delta_f \cos\varphi_s - \varphi_f\delta_s \sin\varphi_s - \varphi_f\delta_f \sin\varphi_s + a\delta_s\varphi_f \sin\varphi_s + a\delta_f\varphi_f \sin\varphi_s.$$

Dropping the terms that vanish after averaging over $T$,

$$\ddot{\delta}_s = \delta_s \cos\varphi_s - a\delta_f \cos\varphi_s - \varphi_f\delta_f \sin\varphi_s + a\delta_s\varphi_f \sin\varphi_s + a\delta_f\varphi_f \sin\varphi_s.$$

Finally, let's remove the higher order terms and average,

$$\ddot{\delta}_s = \delta_s \left(\cos\varphi_s - \overline{\dot{\eta}^2}\cos 2\varphi_s + \overline{\ddot{\eta}\eta^2}\sin^2\varphi_s\cos\varphi_s\right).$$

If we assume that $\varphi_s = \text{const}$ and take into account (12),

$$\ddot{\delta}_s = \delta_s \frac{\sin^2\varphi_s}{\cos\varphi_s}\left(1 + \overline{\ddot{\eta}\eta^2}\cos^2\varphi_s\right). \tag{13}$$

A generic solution is

$$\delta_s = Ae^{-kt} + Be^{kt},$$

where

$$k = \nu_1 = \frac{|\sin\varphi_s|}{\sqrt{\cos\varphi_s}}\left(1 + \overline{\ddot{\eta}\eta^2}\cos^2\varphi_s\right)^{1/2} \tag{14}$$

is the maximal LCE.

We've made several important discoveries. First, from (14), it follows that for $\varphi_s = 0$, the minimum LCE is zero. This means that truly upright stabilization is impossible. The upright equilibrium cannot damp oscillations. It makes intuitive sense: we cannot create any torque by vertical movement if $\varphi = 0$. We can also see it in (11); if $\varphi_s = 0$, then the fast term $\varphi_f$ vanishes.

Second, we can see from (13) that the only way to stabilize the pendulum is to have $\overline{\ddot{\eta}\eta^2} \neq 0$. This can only happen if the pivot motion is asymmetric, that is, the upward motion has a different acceleration profile than the downward motion.

Finally, with strictly periodic motion, when $\overline{\ddot{\eta}\eta^2} = \text{const}$, true stabilization is also impossible. To achieve stabilization and have negative LCE, we must have a damping term in (13),

$$\overline{\ddot{\eta}\eta^2} \sim -\dot{\delta}_s.$$

Therefore, we can expect that if we use the LCE as a reward signal, the agent will essentially go beyond the Kapitza pendulum. To make the LCE negative, it probably will stabilize the pendulum at some small, nonzero angle, and the pivot motion won't be a single cycle periodic motion. In the stationary mode, fast oscillation will be accompanied by slow periodic movement (periodic because of the time symmetry), and in the transition phase, the agent may try to damp the oscillations of $\varphi_s$ by tracking $\dot{\delta}_s$.

## 5 Learning Method

Our environment consisted of the inverted pendulum described by (1). The length of the pendulum rod was 30 cm. ODE (1) was combined with (7) into one system of equations to solve them simultaneously. Note that the immediate output of (7) at time $t$ is useless; we have to see how it evolves through time and use (5) to derive the LCE. To do that, we calculated the maximal LCE according to (5) over a short time window of 0.25 s.

For training, we used Soft Action-Critic (SAC) algorithm [9] from Stable Baselines 3 [10]. We ran 1 s episodes, and each sample in an episode increased time by 5 ms. Action in each sample was acceleration $a$ from (1).

Since the reward in our environment was calculated from the LCE over a time window, the state had to include all values of the pendulum speed and location over the LCE window. That is, our window had 50 time steps, and at each step we recorded $y_c$, $\dot{y}_c$, $\sin\varphi$, $\cos\varphi$, and $\dot{\varphi}$. Therefore, the state was a set of 250 values: $\{(y_{c,i}, \dot{y}_{c,i}, \sin\varphi_i, \cos\varphi_i, \dot{\varphi}_i),\ i = 1\ldots 50\}$, where $i$ is the index of a time step inside the window.

To reset the environment, we set the pendulum at a small random angle within ±5.7° and let it fall for the duration of the LCE time window, filling in all $\varphi$ and $\dot{\varphi}$ in the initial state.

The SAC implementation was used with default parameters, except the learning rate $\alpha = 1 \times 10^{-4}$. The default parameters were: discount $\gamma = 0.99$, soft update rate for target networks $\tau = 5 \times 10^{-3}$, replay buffer size $1 \times 10^6$, batch size 256, train frequency was one gradient step per environment step, target entropy −1, and 100 random steps before training began. The implementation used fully connected neural networks with two hidden layers, 256 neurons each, for all function approximations: actor, critics, and target critics.

We couldn't use pure LCE as a reward signal. The agent would have found a trivial solution: spatially unbounded acceleration down with $a > g$. To encourage the agent to stay in the center, we added a smooth negative reward that was almost constant when $y_c$ stayed

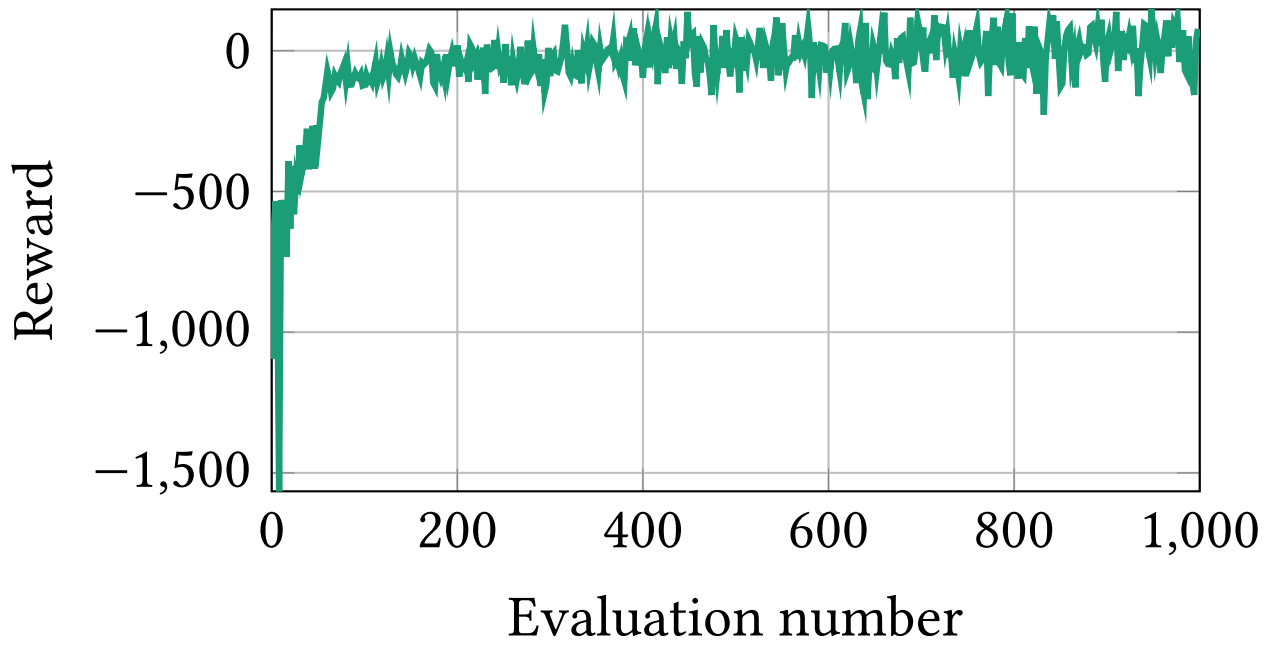

**Figure 2:** Training progress.

within $\pm y_{\max}$ and rapidly increased when the agent went outside these bounds:

$$r_t = -v_1 - 100\frac{(y_c/y_{\max})^{20}}{1+(y_c/y_{\max})^{20}}. \tag{15}$$

If the agent went outside the $\pm y_{\max}$ bounds, we terminated the episode with $r_t = -1000$. We had $y_{\max} = 10\,\text{cm}$ in our experiments.

## 6 Results

Figure 2 shows the training progress. The quick rise indicates that the agent rapidly learned how to stay within the $\pm y_{\max}$ boundary. Then it slowly learned how to stabilize the pendulum.

The work of the trained agent is shown in fig. 3a. The agent's performance is remarkable. Not only did it find a mode that provides an equilibrium point at the top, it damped the angular oscillations, significantly improving upon the Kapitza pendulum. For comparison, the same picture shows how the Kapitza pendulum behaves when we oscillate the pivot point with the same frequency as the trained model. The Kapitza pendulum started from 1° and continued oscillating. The agent started from 10° with the pendulum falling. It stopped the fall at 15° and then brought the pendulum to oscillations as small as ±0.01° (see fig. 3d).

Note that training episodes were 2 s. In this time span, the pendulum has not yet stabilized. At the end of the episode the agent was still damping the oscillations. However, the learned policy was general enough to let the agent keep the pendulum upright outside the training window.

Since in the previous work [3] the authors used the DDPG learning algorithm, we verified whether SAC would stabilize the inverted pendulum with the same reward signal as they used. We changed $v_1$ in (15) to $c\varphi^2$, and $c = 25$ was selected to keep the reward roughly the same order of magnitude as the LCE. The result is shown in fig. 4. We can see that the agent managed to keep the pendulum upright, but the pendulum moves chaotically around the equilibrium with a far larger amplitude than the pendulum driven by an agent that was trained on the LCE signal.

## 7 Conclusion

We suggested a physics-informed reward signal based on the LCE in an attempt to let the agent discover the Kapitza pendulum. The trained model outperformed it. Not only did it find an equilibrium point at the top, it also managed to damp the oscillations and bring the pendulum to almost a halt with residual oscillations of ±0.01°.

The remaining question is why the stable mode keeps the pendulum at 0.25° from vertical. Is it a fundamental constant that minimizes the LCE? Can we make it zero? Can we stabilize the pendulum at an arbitrary angle from vertical?

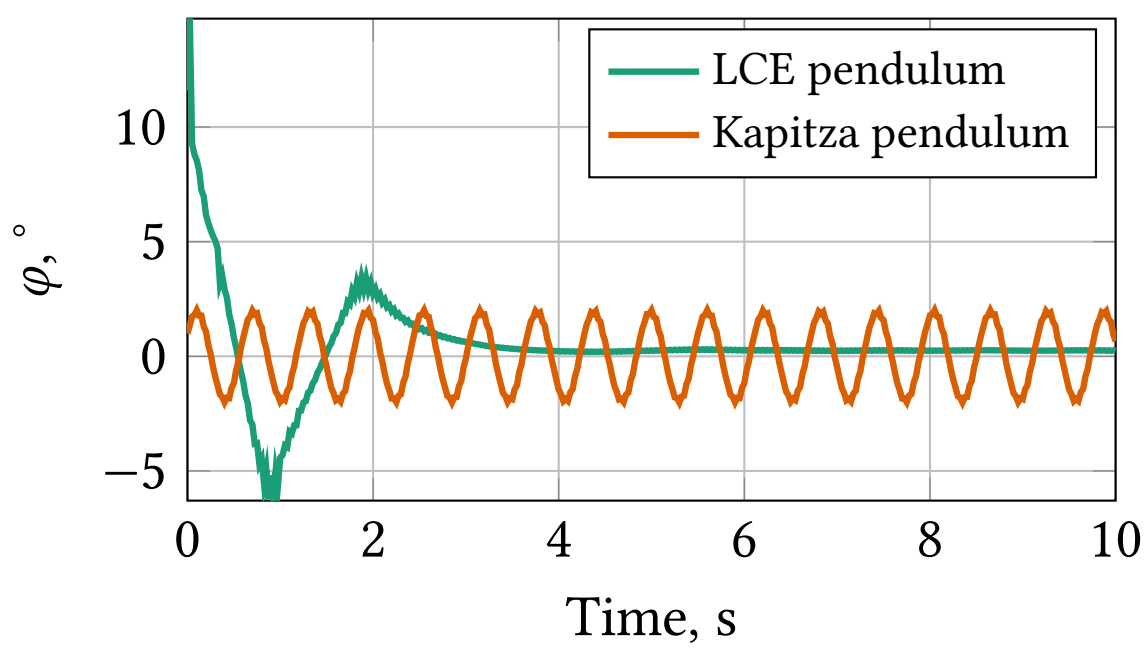

**(a)** Vertical angle for the LCE pendulum and the Kapitza pendulum for comparison. The initial position of the latter was 1°.

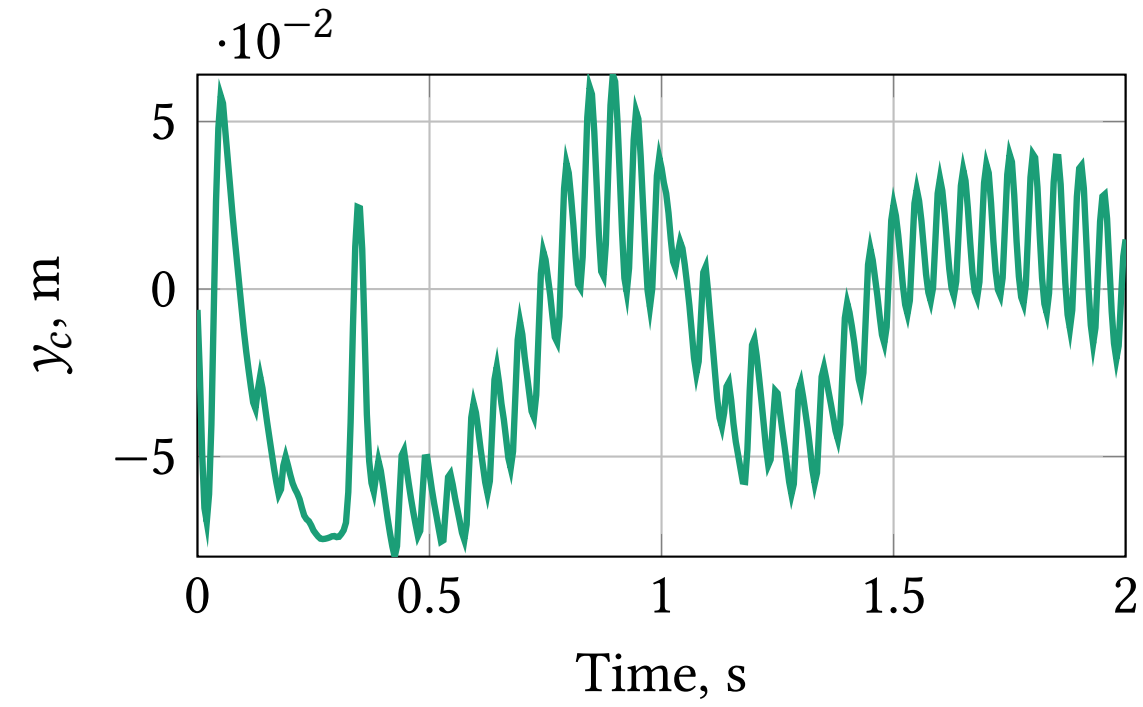

**(b)** The agent is damping oscillations at the beginning of an episode.

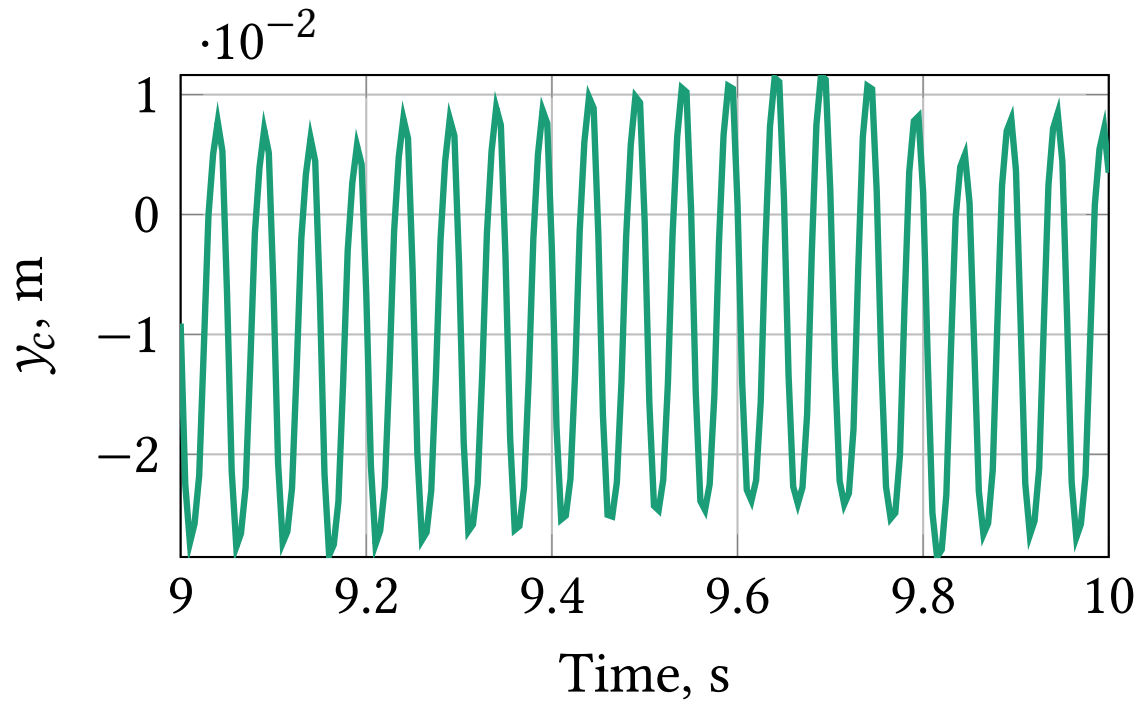

**(c)** Pivot point vertical motion in stable mode.

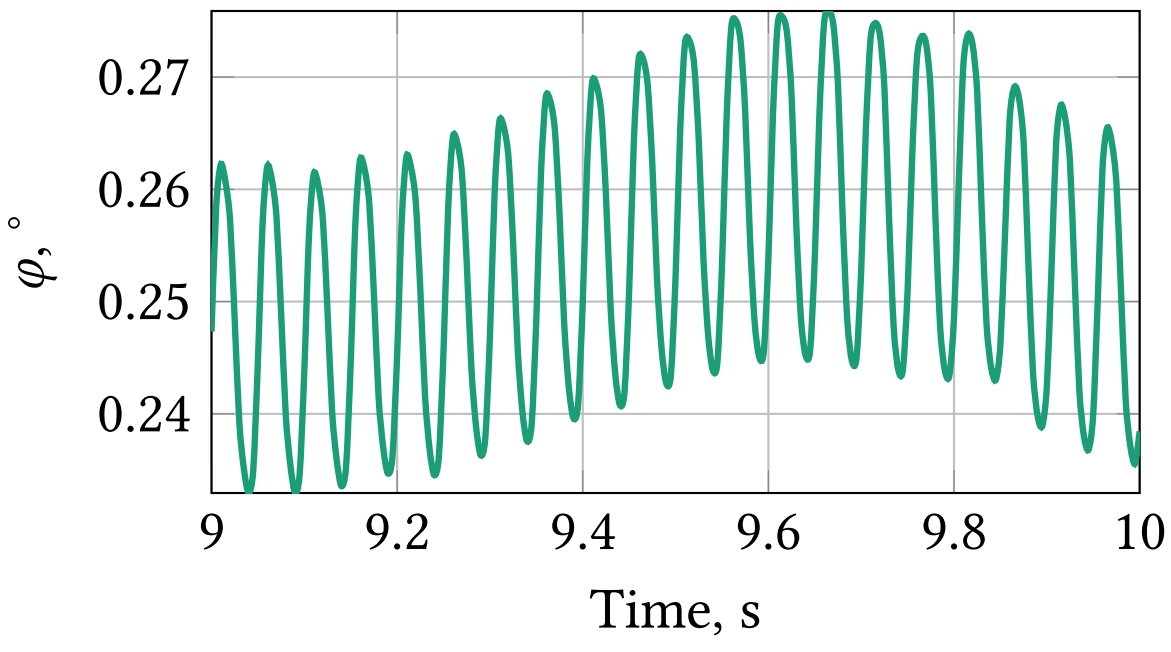

**(d)** Residual $\varphi$ oscillations in stable mode.

**Figure 3:** Inverted pendulum stabilization with a trained agent.

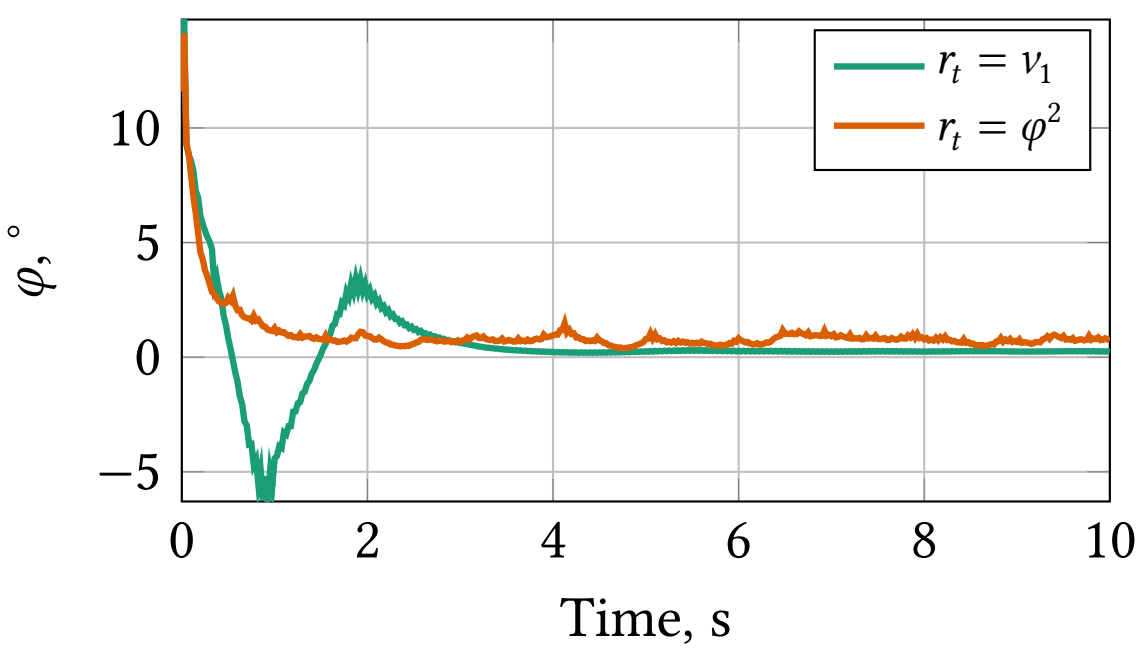

**Figure 4:** Comparison of models trained with the LCE and $\varphi^2$ reward signals.